\PassOptionsToPackage{table}{xcolor}
\documentclass{article} 
\usepackage{iclr2027_conference,times}

\usepackage{amsmath,amsfonts,bm}

\def\eqref#1{equation~\ref{#1}}

\def\1{\bm{1}}

\DeclareMathAlphabet{\mathsfit}{\encodingdefault}{\sfdefault}{m}{sl}
\SetMathAlphabet{\mathsfit}{bold}{\encodingdefault}{\sfdefault}{bx}{n}

\usepackage{booktabs}          
\usepackage{multirow}          
\usepackage{xcolor}            
\usepackage{makecell}          
\usepackage{hyperref}
\usepackage{url}
\usepackage{fontawesome5}      
\usepackage{bbm}
\usepackage{amssymb}          
\usepackage{subcaption}
\usepackage{wrapfig}
\usepackage{graphicx}
\usepackage{algorithm}
\usepackage{algpseudocode}

\definecolor{badgegreen}{RGB}{56,142,60}  

\title{Reinforcement Learning with Visually Attributable On-Policy Self-Distillation}

\author{Shiao Xie\thanks{Shiao Xie and Siyu Chen contributed equally to this work.}, Siyu Chen\footnotemark[1]\textsuperscript{\enspace}\thanks{Siyu Chen and Yujin Wang are the corresponding authors.}, Jianwei Lv, Xiandong Li, Yujin Wang\footnotemark[2] \\
Baidu Inc., China
}

\newlength{\vagtabh}

\iclrfinalcopy 
\begin{document}

\maketitle

\begin{center}
\large
\href{https://huggingface.co/siyu-ch/VAG-Qwen3.5-4B}{\textcolor{badgegreen}{\faCube~Qwen3.5-4B}}\quad
\href{https://huggingface.co/siyu-ch/VAG-Qwen3.5-9B}{\textcolor{badgegreen}{\faCube~Qwen3.5-9B}}
\end{center}

\begin{abstract}
Multimodal reinforcement learning with verifiable rewards (RLVR) provides reliable task-level supervision, yet offers limited guidance for localizing errors in visual reasoning.
On-policy self-distillation (OPSD) complements this supervision with dense token-level corrections from a privileged teacher equipped with an evidence-centered visual view.
However, the resulting corrections may conflate visual evidence with non-visual shifts induced by linguistic priors, response style, or the privileged view itself.
We formalize this phenomenon as \emph{privileged-correction contamination}: a token-level correction is contaminated when its update direction cannot be consistently attributed to the privileged visual evidence.
Directly distilling contaminated corrections may therefore reinforce erroneous token-level updates.
To address this issue, we propose reinforcement learning with visually attributable on-policy self-distillation, an evidence-consistent paradigm that augments task-level RLVR supervision with privileged token-level corrections grounded in visual evidence.
We instantiate this paradigm with \textbf{Visual-Attribution Gating (VAG)}, which uses the verifier-derived advantage as the optimization anchor and validates each privileged correction against the contrast between evidence-present and evidence-degraded teacher views.
By admitting only corrections directionally supported by this visual contrast, VAG preserves visually grounded guidance while suppressing inconsistent residuals.
Experiments across six visual reasoning benchmarks show that VAG consistently outperforms GRPO and OPSD methods, establishing new state-of-the-art average performance. 

\end{abstract}

\section{Introduction}
\label{intro}
Reinforcement learning with verifiable rewards (RLVR) has become an effective
paradigm for improving multimodal reasoning by optimizing policies against
automatically checkable outcomes~\citep{guo2025deepseek, yang2025qwen3}. 
In GRPO~\citep{grpo}, a verifier scores each rollout and group-relative
normalization turns these rewards into an outcome advantage $A_{\mathrm{ORM}}$
shared by every token. Such supervision is
reliable at the trajectory level, but coarse at the token level: the same
advantage is assigned to every token, regardless of which visual decision
determines the answer. Consequently, RLVR can distinguish successful
trajectories from unsuccessful ones without localizing the visual reasoning
errors within them.

On-policy distillation (OPD)~\citep{agarwal2024onpolicy, gu2023minillm, song2026survey} addresses this limitation by evaluating
student-sampled trajectories with a teacher and supplying dense token-level
guidance. On-policy self-distillation (OPSD)~\citep{zhao2026opsd, hubotter2026sdpo, rlsd2026}
further avoids dependence on an external teacher by using the policy itself
under privileged information as the teacher, which for multimodal reasoning can
be a zoomed-in view that exposes task-relevant evidence~\citep{visionopd}.
The resulting teacher--student log-probability gap $\delta_t$ provides a dense,
token-level correction, yet it is not guaranteed to be attributable to the
privileged visual evidence alone: it may also be contaminated by non-visual
shifts induced by linguistic priors, response style, or the privileged view
itself.

We formalize this contamination as \emph{privileged-correction contamination}: the
teacher--student correction mixes evidence-grounded guidance with residual
effects that are not attributable to the visual evidence. 
Applying such a correction directly may reinforce a token even
when the visual evidence argues against it, or suppress a token that the
evidence supports. 

\begin{wrapfigure}{r}{0.57\columnwidth}
    \centering
    \vspace{-0.8\baselineskip}
    \includegraphics[width=\linewidth]{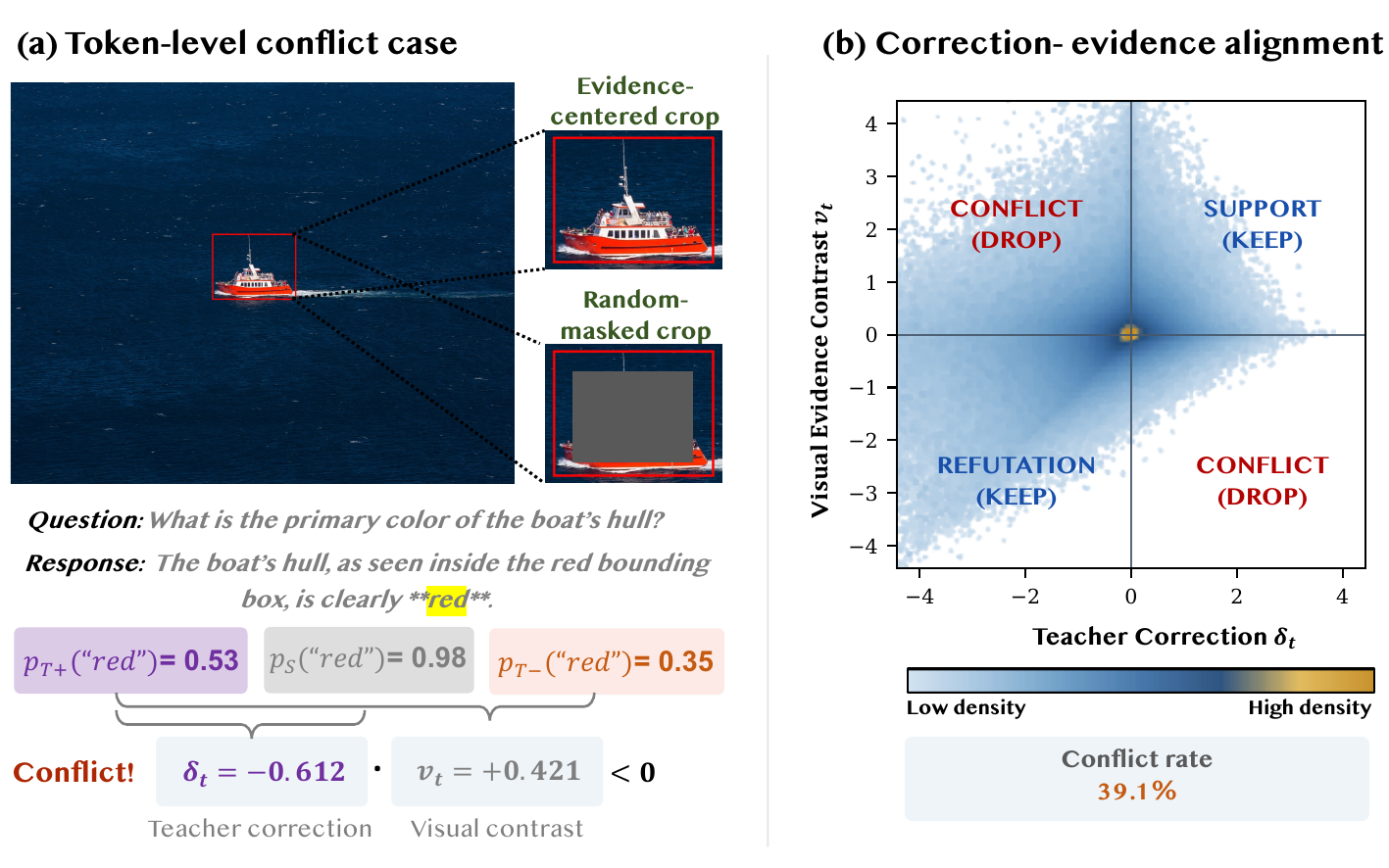}
    \caption{Illustrative contamination diagnostic. (a) A token-level
    conflict case and (b) the correction-evidence alignment used to motivate attribution.}
    \label{fig:contamination}
    \vspace{-0.6\baselineskip}
\end{wrapfigure}

To make this contamination concrete, Fig.~\ref{fig:contamination} illustrates a single token-level diagnostic.
For each generated token, we compare the teacher correction $\delta_t$ with
the visual contrast $v_t$ between evidence-centered and random-masked teacher
views, a probe of visual dependence introduced by OPD-V~\citep{opdv}.
Fig.~\ref{fig:contamination}~(a) shows a conflicting case, in
which the teacher suppresses a token ($\delta_t=-0.612$) that the visual contrast supports ($v_t=+0.421$).
Such a correction should be discarded rather than applied,
since the student already assigns the token a higher
probability than the teacher does ($0.98$ vs. $0.53$): the teacher is lower
here because the privileged view shifts its probabilities downward, not because
the visual evidence refutes the token.
Fig.~\ref{fig:contamination}~(b) summarizes the comparison over tokens
by the signs of the two signals:
agreeing signs indicate a correction consistent with the evidence,
whereas opposite signs expose the two conflict cases, in which the teacher
suppresses a visually supported token or promotes a visually refuted one. Measured with
Qwen3.5-4B on our training data (\S\ref{sec:implementation-details}), these
conflicts affect $39.1\%$ of the generated tokens.
Exposing these conflicts, however, does not remove them. What the diagnostic
calls for is visual attribution in a strict sense: an explicit mechanism that
checks whether the \emph{direction} of each correction agrees with the visual
evidence.

Motivated by this diagnosis, we design \textbf{Visual-Attribution Gating (VAG)}. 
Rather
than distilling every privileged correction, VAG validates the direction of each
one against the visual contrast: the correction $\delta_t$ is admitted only when 
it agrees in sign with $v_t$, that is $v_t \delta_t>0$, 
which keeps both supporting and refuting corrections and removes
only the conflicting residuals exposed by the diagnostic.
Furthermore, VAG operates on top of RLVR rather than in place of it: the
verifier-derived advantage remains the optimization anchor, and self-distillation
contributes only the corrections that pass the gate. 
Taken
together, we propose reinforcement learning with visually attributable
on-policy self-distillation, a new evidence-consistent learning paradigm.

\begin{figure*}[t]
    \centering
    \includegraphics[width=\textwidth]{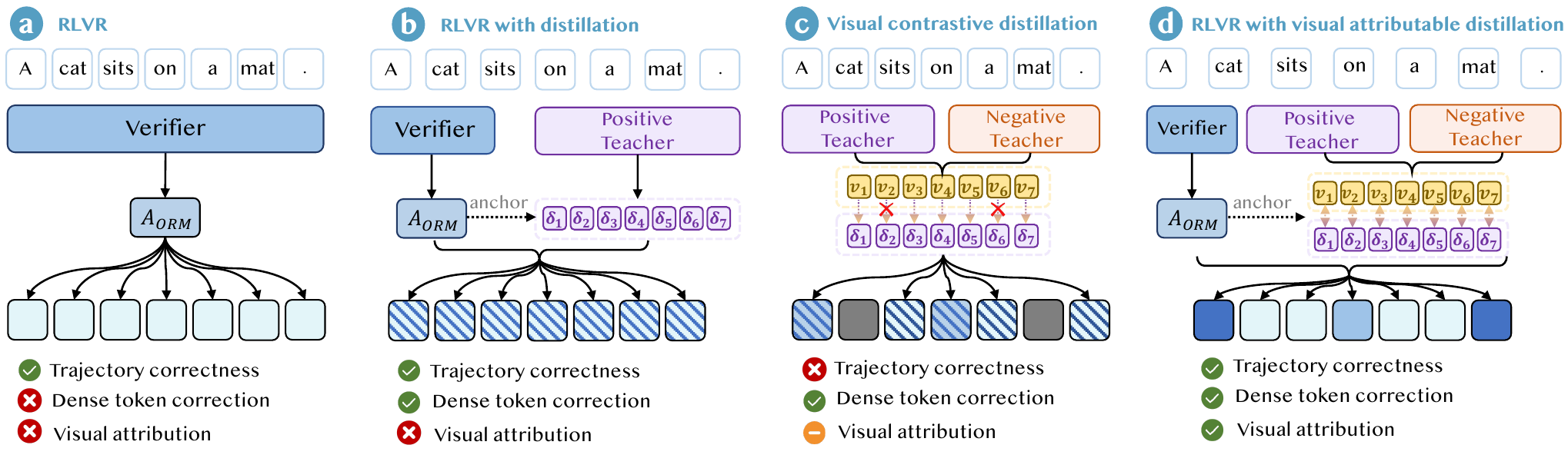}
    \caption{
\textbf{Comparison of on-policy learning paradigms.} 
 (a) RLVR, (b) RLVR with distillation, (c) Visual contrastive distillation, and (d) RLVR with visual attributable distillation.
    }
    \label{fig:motivation}
\end{figure*}

Fig.~\ref{fig:motivation} compares our paradigm with representative
on-policy learning paradigms for multimodal credit assignment
along three properties: \emph{trajectory correctness},
\emph{dense token correction}, and \emph{visual attribution}.
RLVR (Fig.~\ref{fig:motivation}~(a)) supplies only the outcome advantage
$A_{\mathrm{ORM}}$, which anchors optimization to trajectory correctness but
assigns the same scalar to every token.
RLVR with distillation (Fig.~\ref{fig:motivation}~(b)) keeps this anchor and
modulates each token by $\delta_t$ without verifying where that correction comes
from, so the privileged signal it distills may be
contaminated~\citep{opdvr2026, rlsd2026}.
Visual contrastive distillation (Fig.~\ref{fig:motivation}~(c)) instead uses
visual contrast $v_t$ to select only visually sensitive tokens~\citep{opdv}. 
Yet this selection does not verify whether an update follows the visual
evidence, so the corrections applied to the selected tokens remain contaminated,
while the verifier-grounded anchor is lost as well.

VAG (Fig.~\ref{fig:motivation}~(d)) is the only paradigm that attains all three
properties at once: it keeps $A_{\mathrm{ORM}}$ as the anchor, preserves the
dense correction $\delta_t$, and admits that correction only when $v_t$ supports
its direction. This coverage comes at almost no cost. VAG needs no learned
contamination estimator and no auxiliary distillation objective, and relative to
a dual-view OPSD baseline that already evaluates both teacher views, it adds only
an elementwise sign test and no additional teacher forward pass. 
Across six visual reasoning benchmarks, VAG consistently improves over matched
GRPO and OPSD baselines and surpasses much larger open-source models (e.g., Qwen3.5-397B), leading
closed-source models (e.g., GPT-5.4), and agentic ``Thinking-with-Images'' methods on tasks that
require dense local evidence. 
On holdout tasks beyond the training distribution,
it retains general visual understanding and reasoning ability, indicating that
the gains do not come at the cost of forgetting, while controlled ablations
isolate the contribution of the sign-agreement gate.

Our contributions are summarized as follows:
\begin{itemize}
    \item We identify \emph{privileged-correction contamination}, 
    showing that dense teacher–student corrections are not necessarily aligned in direction with the visual evidence.
    \item We propose an evidence-consistent paradigm unifying RLVR and OPSD, in
    which the verifier-derived advantage anchors optimization, and design
    VAG to mitigate privileged-correction contamination by
    admitting a teacher correction only under sign agreement with a controlled
    visual contrast.
    \item Across six visual reasoning benchmarks, VAG consistently improves over
    matched GRPO and OPSD baselines and surpasses much larger open-source and
    leading closed-source models, while holdout evaluation shows no loss of
    general capability. Ablations further validate the proposed attribution
    mechanism.
\end{itemize}

\section{Related Work}

\subsection{On-Policy and Privileged Distillation}
OPD supervises student-visited prefixes with a
teacher, thereby reducing exposure bias while supplying dense token-level
guidance~\citep{gu2023minillm,agarwal2024onpolicy,song2026survey}.
OPSD replaces the external teacher with the model
itself conditioned on privileged context, such as verified solutions, reasoning
traces, or environmental feedback
~\citep{zhao2026opsd,hubotter2026sdpo,he2026self,jin2026unisd}. Because the
student cannot access this context at inference time, the resulting
teacher--student gap may conflate task-relevant knowledge with leakage or
stylistic artifacts~\citep{rlsd2026,rlcsd2026,shen2026anti}. In multimodal
reasoning, privileged views instead expose localized evidence that global image
processing tends to lose, and methods from region-to-image distillation to
Vision-OPD transfer such evidence to a full-image
model~\citep{wei2026zooming,visionopd}. Subsequent work refines the selectivity
of this supervision through view-based reweighting~\citep{vaopd}, trajectory
gating~\citep{vzero}, gradient decomposition~\citep{decomposedopd}, modality
balance~\citep{opdv}, and target reconstruction~\citep{vad2026}, whereas VAG
uses a signed visual contrast to determine whether a candidate teacher
correction is admitted into policy optimization at all.

\subsection{Integrating RLVR with On-Policy Distillation}
RLVR anchors optimization to checkable outcomes, and GRPO realizes this without a learned
critic by estimating group-relative advantages that are nevertheless shared by
all tokens in a response~\citep{grpo}. OPD supplies the missing token-level
resolution, yet its preference need not agree with trajectory
correctness, so the two signals are complementary but not automatically
compatible. Existing integrations reconcile them by weighting the two
objectives, restricting the teacher gap to modulating verifier-directed
magnitudes, gating implicit OPD rewards by trajectory correctness, or
contrasting privileged contexts
~\citep{hubotter2026sdpo,rlsd2026,opdvr2026,rlcsd2026}, while  variants
route samples or tokens between the two objectives, account for student
uncertainty, or reverse teacher signals to encourage exploration
~\citep{li2026unifying,wang2026trace,ke2026respecting,kim2026rebellious}. VAG
resolves this compatibility at the token level instead: a signed correction
enters the verifier-anchored advantage only when its direction agrees with the
visual evidence contrast, which preserves both visually supported promotion and
visually justified suppression while falling back to outcome supervision
whenever attribution fails.

\section{Method}
\label{sec:method}



\begin{figure*}[t]
    \centering
    \includegraphics[width=1.\textwidth]{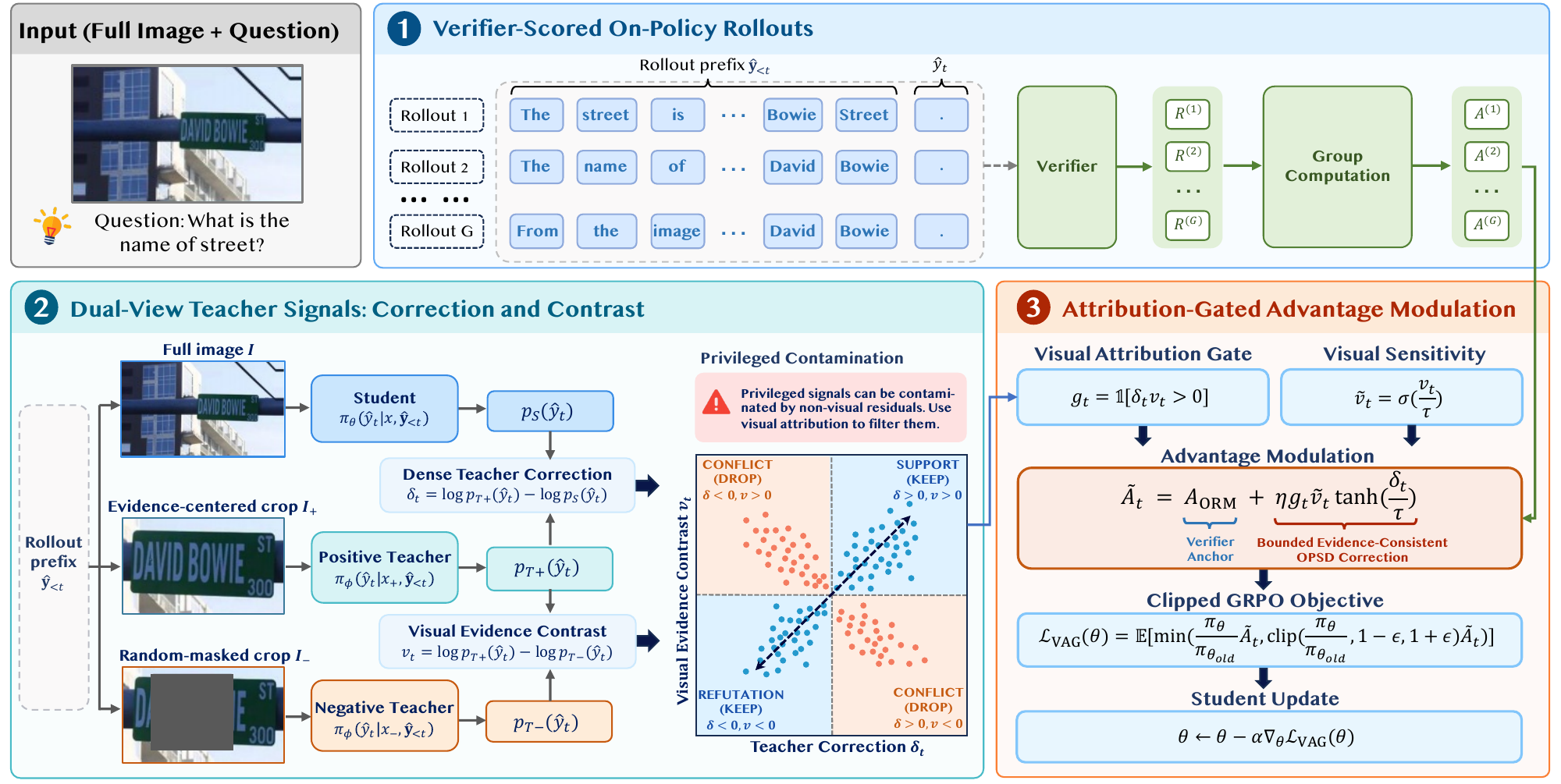}
    \caption{
    Overview of the VAG framework.
    \textbf{Stage 1:} A verifier derives $A_{\mathrm{ORM}}$ from on-policy
    rollouts. \textbf{Stage 2:} Dual-view teacher evaluation yields the
    distillation margin $\delta_t$ and visual contrast $v_t$; VAG retains same-sign
    corrections and discards conflicting ones. \textbf{Stage 3:} The retained
    correction modulates $A_{\mathrm{ORM}}$ for the clipped GRPO update.
    }
    \label{fig:vag_framework}
\end{figure*}

\subsection{Overview}
\label{sec:overview}
Given an image–question pair, VAG proceeds through the three-stage training
loop illustrated in Fig.~\ref{fig:vag_framework}. In Stage 1
(Sec.~\ref{sec:rollout-verifier}), VAG derives a reliable trajectory-level
advantage from on-policy rollouts sampled by the frozen student policy and
scored by a task verifier. In Stage 2
(Sec.~\ref{sec:visual-opsd}), the positive and negative teacher branches of
the same Exponential Moving Average (EMA) model replay each student-generated prefix under an
evidence-centered view and its random-masked counterpart, respectively. These
   evaluations yield a teacher–student correction $\delta_t$ and a visual contrast
   $v_t$, whose sign agreement retains evidence-consistent token-level
supervision while filtering potentially contaminated corrections.
In Stage 3 (Sec.~\ref{sec:advantage-modulation}), VAG weights each retained
correction by its visual sensitivity and uses it to modulate, rather than
replace, the verifier advantage. The resulting token-level advantage is then
optimized using the clipped GRPO objective.

\subsection{Verifier-Scored On-Policy Rollouts}
\label{sec:rollout-verifier}
Let $q$ denote a question and $I$ its corresponding full image, with the
multimodal input denoted by $x=(q,I)$. At each training iteration,
$\pi_{\theta}$ denotes the student policy used for rollout collection. As shown in Stage 1 of
Fig.~\ref{fig:vag_framework}, the student samples a group of $G$ response
trajectories as:
\begin{equation}
\label{eq:on-policy-rollout}
    \hat{\mathbf{y}}^{(i)}
    =
\bigl(\hat{y}^{(i)}_1,\ldots,\hat{y}^{(i)}_{T_i}\bigr)
    \sim
    \pi_{\theta}(\cdot \mid x),
    \qquad i=1,\ldots,G,
\end{equation}
where $\hat{\mathbf{y}}^{(i)}$ is the $i$-th sampled response,
$\hat{y}^{(i)}_t$ is its token at position $t$, and $T_i$ is its length. A task
verifier assigns each rollout an outcome reward $R^{(i)}\in \left \{ 0,1 \right \}$, from which we
compute the group-normalized advantage as: 
\begin{equation}\label{eq:aorm}
  A_{\mathrm{ORM}}^{(i)}
  = \frac{R^{(i)}-\mu_G}{\sigma_G},
\end{equation}
where $\mu_G$ and $\sigma_G$ denote the mean and standard deviation of the
rewards within the rollout group, respectively. The scalar
$A_{\mathrm{ORM}}^{(i)}$ is shared by all tokens in rollout $i$. Although it
provides a reliable trajectory-level learning signal, it does not identify
which tokens contribute to the outcome.




\subsection{Dual-View Teacher Signals: Correction and Contrast}
\label{sec:visual-opsd}
Stage~1 provides a reliable outcome-level anchor but leaves token-level credit
undetermined. OPSD closes this gap by letting a privileged teacher re-score the
student's own rollout token by
token~\citep{zhao2026opsd}, and we adopt
this mechanism as our dense signal, instantiating the teacher with an
evidence-centered view. 
Unlike prior OPSD, which distills every
correction directly, we first ask whether a correction is directionally
supported by the visual evidence, following the diagnostic in Sec.~\ref{intro}.
As
shown in Stage 2 of Fig.~\ref{fig:vag_framework}, both teachers are
view-conditioned branches of a single EMA teacher $\pi_{\phi}$ replaying the
student's trajectory: conditioning $\pi_{\phi}$ on the evidence-preserving view
yields the \emph{positive teacher}, and conditioning it on the
evidence-degraded view yields the \emph{negative teacher}. 
Because the two
branches share the student-generated prefix and differ only in the visual
condition, all token-level quantities below are comparable.
This
yields two steps: form the privileged correction from the positive teacher, then
probe its visual support with the negative teacher.

\paragraph{Dense Teacher Correction.}
The first step instantiates the positive branch. The evidence-preserving view
$I_{+}$ is a crop centered on the target detail relevant to the question, which
gives the positive teacher input $x_{+}=(q,I_{+})$. For each student-generated
trajectory $\hat{\mathbf{y}}^{(i)}$, the positive teacher re-evaluates every
generated token $\hat{y}^{(i)}_t$ conditioned on the same student-generated
prefix $\hat{\mathbf{y}}^{(i)}_{<t}
=(\hat{y}^{(i)}_1,\ldots,\hat{y}^{(i)}_{t-1})$. We write the corresponding
sampled-token probabilities as:
\begin{align}
\label{eq:token-probs}
    {p_{S}^{(i)}}(\hat{y}_t)
    &\triangleq
     \pi_{\theta}\!\left(
        \hat{y}^{(i)}_t
        \mid x,\hat{\mathbf{y}}^{(i)}_{<t}
    \right),  
    \\
    \label{eq:token-probs-pos}
    {p_{T+}^{(i)}}(\hat{y}_t)
    &\triangleq
    \pi_{\phi}\!\left(
        \hat{y}^{(i)}_t
        \mid x_{+},\hat{\mathbf{y}}^{(i)}_{<t}
    \right),
\end{align}
for the student and the positive teacher, respectively. The dense teacher
correction is their log-ratio on the sampled token:
\begin{equation}
\label{eq:dt}
    \delta_t^{(i)}
    =
    \operatorname{sg}\!\left[
        \log {p_{T+}^{(i)}}(\hat{y}_t)-\log{p_{S}^{(i)}}(\hat{y}_t)
    \right],
\end{equation}
where $\operatorname{sg}[\cdot]$ denotes the stop-gradient operator. A positive
$\delta_t^{(i)}$ indicates that the positive teacher assigns the sampled token a
higher probability than the student, and a negative value the opposite, so
$\delta_t^{(i)}$ is precisely the dense signal that on-policy self-distillation
would follow.

Taken on its own, however, $\delta_t^{(i)}$ is exactly the quantity examined in
Fig.~\ref{fig:contamination}: a non-zero teacher--student gap cannot be assumed
to follow the direction of the privileged evidence. We therefore treat
$\delta_t^{(i)}$ as a candidate correction and, instead of assuming its visual
support, make that support testable.

\paragraph{Visual Evidence Contrast.}
Following OPD-V~\citep{opdv}, we pair $I_{+}$ with an evidence-degraded
counterpart $I_{-}$, obtained by occluding a randomly placed rectangular region
of the zoom-in image with black pixels, so that the visual evidence available to
the teacher is weakened while the image resolution, the visual token grid, and
the surrounding spatial context remain unchanged. 
Feeding
$x_{-}=(q,I_{-})$ to the same teacher on the same prefix gives: 
\begin{equation}
\label{eq:token-probs-neg}
    {p_{T-}^{(i)}}(\hat{y}_t)
    \triangleq
    \pi_{\phi}\!\left(
        \hat{y}^{(i)}_t
        \mid x_{-},\hat{\mathbf{y}}^{(i)}_{<t}
    \right),
\end{equation}
and because the two teacher evaluations share parameters, prefix, and realized
token, and differ only in the target evidence, their log-ratio isolates the
teacher's response to that intervention:
\begin{equation}
\label{eq:st}
    v_t^{(i)}
    =
    \operatorname{sg}\!\left[
        \log {p_{T+}^{(i)}}(\hat{y}_t)-\log {p_{T-}^{(i)}}(\hat{y}_t)
    \right].
\end{equation}
A positive $v_t^{(i)}$ indicates that the target evidence supports the sampled
token, whereas a negative value indicates refutation, and $\lvert
v_t^{(i)}\rvert$ measures how strongly the token depends on that evidence.

\paragraph{Diagnosing Privileged Contamination.}
Pairing the two signals by sign partitions the generated tokens into the four
quadrants shown in Stage~2 of Fig.~\ref{fig:vag_framework}. The two same-sign
quadrants are directionally consistent: when $\delta_t>0$ and $v_t>0$, the
teacher promotes a token whose probability the visual evidence also raises
(\emph{support}); when $\delta_t<0$ and $v_t<0$, the teacher suppresses a token
whose probability the evidence lowers (\emph{refutation}). In both cases the
candidate correction moves the student in the same direction as the visual
intervention, so it can be treated as evidence-grounded credit. Notably, keeping
the refutation quadrant is what allows suppressive corrections to survive rather
than be discarded together with the conflicting ones.

The two opposite-sign quadrants are directionally conflicting. A correction with
$\delta_t>0$ but $v_t<0$ would reinforce a token whose probability the visual
evidence lowers, whereas $\delta_t<0$ with $v_t>0$ would suppress a token whose
probability it raises. Since neither direction can be explained by the visual
intervention, we regard these quadrants as observable instances of
privileged-correction contamination and exclude them from the dense supervision.
The four quadrants thus define a binary admissibility rule on candidate
corrections, and what remains is to realize this rule as a token-level learning
signal without discarding the verifier anchor of Stage~1.

\subsection{Attribution-Gated Advantage Modulation}
\label{sec:advantage-modulation}
Stage~3 realizes the admissibility rule as a token-level advantage: the gate
decides \emph{whether} a candidate correction enters the update, the contrast
magnitude decides \emph{how strongly} it enters, and the outcome advantage of
Stage~1 remains the anchor that determines the direction of the update. 
The visual attribution gate shown in Fig.~\ref{fig:vag_framework} is formulated as:
\begin{equation}\label{eq:mask}
g_t^{(i)}=\mathbbm{1}\!\left[\delta_t^{(i)}v_t^{(i)}>0\right],
\end{equation}
where $\mathbbm{1}[\cdot]$ is the indicator function. This gate retains the
support and refutation quadrants and rejects both conflict quadrants, together
with the boundary cases that carry no directional information.
We therefore convert the
contrast into a bounded visual-sensitivity weight:
\begin{equation}\label{eq:gate}
  \widetilde{v}_t^{(i)}=\sigma\!\left(
  \frac{v_t^{(i)}}{\tau}\right)\in(0,1),
\end{equation}
where $\sigma$ is the logistic sigmoid and $\tau>0$ controls the sharpness of
the transition, 
so that strongly evidence-dependent tokens receive more credit
than marginally dependent ones instead of being weighted equally. The teacher
correction is bounded in the same spirit, and the two are combined with the
verifier anchor to form the modulated advantage:
\begin{equation}\label{eq:mod-adv}
  \widetilde{A}_t^{(i)}
  =A_{\mathrm{ORM}}^{(i)}
  +\eta\,g_t^{(i)}\widetilde{v}_t^{(i)}
  \tanh\!\left(\frac{\delta_t^{(i)}}{\tau}\right),
\end{equation}
where $\eta>0$ controls the strength of the correction. The first term provides
a trajectory-level verifier anchor, while the second redistributes credit at
the token level only when the teacher correction is visually attributable. If
$g_t^{(i)}=0$, the correction vanishes and
$\widetilde{A}_t^{(i)}=A_{\mathrm{ORM}}^{(i)}$, so filtering a 
correction falls back to outcome supervision rather than removing the token from
training.

The final stage of Fig.~\ref{fig:vag_framework} optimizes the student policy
using the modulated advantage. We first define the token-level importance
ratio between the current policy $\pi_\theta$ and the rollout policy
$\pi_{\theta_{old}}$ as:
\begin{equation}\label{eq:rho}
  \rho_t^{(i)}(\theta)
  = \frac{\pi_\theta(\hat{y}_t^{(i)} \mid x, \hat{\mathbf{y}}_{<t}^{(i)})}
         {\pi_{\theta_{old}}(\hat{y}_t^{(i)} \mid x, \hat{\mathbf{y}}_{<t}^{(i)})}. 
\end{equation}
The training objective is the clipped GRPO surrogate averaged over rollouts
and token positions:
\begin{equation}\label{eq:loss}
  \mathcal{L}_{\mathrm{VAG}}(\theta)
  = -\frac{1}{G}\sum_{i=1}^{G}\frac{1}{T_i}\sum_{t=1}^{T_i}
    \min\!\Big(
      \rho_t^{(i)}(\theta)\,\widetilde{A}_t^{(i)},\;
      \operatorname{clip}\big(
        \rho_t^{(i)}(\theta),\,1-\varepsilon,\,1+\varepsilon
      \big)\,\widetilde{A}_t^{(i)}
    \Big),
\end{equation}
where $0<\varepsilon<1$ is the clipping radius. All teacher-derived signals,
including $\delta_t$, $v_t$, and $g_t$, are detached, so
gradients flow to the student only through $\rho_t^{(i)}(\theta)$. 
Consequently, VAG introduces neither additional trainable parameters nor an
auxiliary objective; it operates entirely by reshaping the advantage that GRPO
already optimizes.


\begin{table*}[t]
\centering
\small
\caption{
Comparison with single-forward-pass MLLMs and 
post-training methods across multimodal reasoning benchmarks.
All entries report accuracy (\%).
}
\setlength{\tabcolsep}{2.6pt}
\renewcommand{\arraystretch}{1.1}
\resizebox{\textwidth}{!}{
\begin{tabular}{l c c c c c c c c}
\toprule[1.5pt]
\textbf{Model}
& \textbf{Param.}
& \textbf{V*}
& \textbf{Zoom}
& \makecell{\textbf{HR-Bench}\\\textbf{4K}}
& \makecell{\textbf{HR-Bench}\\\textbf{8K}}
& \makecell{\textbf{MME-RW}\\\textbf{EN}}
& \makecell{\textbf{MME-RW}\\\textbf{CN}}
& \textbf{Avg.} \\
\midrule

\multicolumn{9}{c}{
    \textit{Closed-Source Models (Single Forward Pass)}
} \\
\midrule
GPT-5.2 {\scriptsize\citep{gpt52}}
& -- & 79.06 & 50.89 & 81.12 & 78.38 & 72.60 & 68.80 & 71.81 \\

GPT-5.4 {\scriptsize\citep{gpt54}}
& -- & 76.96 & 52.66 & 84.00 & 77.88 & 74.20 & 70.93 & 72.77 \\

Gemini 3.1 Pro {\scriptsize\citep{gemini31pro}}
& -- & 87.96 & 61.18 & 89.63 & 86.88 & 76.53 & 73.31 & 79.25 \\

Gemini 3.5 Flash {\scriptsize\citep{gemini35flash}}
& -- & 89.01 & 61.42 & 89.12 & 86.62 & 75.31 & 73.97 & 79.24 \\

\midrule
\multicolumn{9}{c}{
    \textit{Open-Source Baselines (Single Forward Pass)}
} \\
\midrule
DeepEyes {\scriptsize\citep{deepeyes}}
& 7B & 85.86 & 46.51 & 75.13 & 72.63 & 64.10 & 64.09 & 68.05 \\

Thyme {\scriptsize\citep{thyme}}
& 7B & 82.20 & 45.09 & 77.00 & 72.00 & 64.80 & 64.59 & 67.61 \\

DeepEyesV2 {\scriptsize\citep{deepeyesv2}}
& 7B & 81.68 & 44.97 & 77.88 & 73.75 & 64.90 & 65.07 & 68.04 \\

SenseNova-MARS {\scriptsize\citep{sensenovamars}}
& 8B & 92.15 & 47.81 & 83.13 & 78.38 & 67.90 & 68.90 & 73.05 \\

MiMo-VL-RL {\scriptsize\citep{mimovl}}
& 7B & 83.25 & 45.68 & 73.50 & 69.38 & 62.73 & 55.89 & 65.07 \\

ZwZ {\scriptsize\citep{wei2026zooming}}
& 8B & 87.96 & 56.69 & 83.63 & 81.75 & 66.57 & 68.09 & 74.12 \\

MiniCPM-V-4.5 {\scriptsize\citep{minicpmv45}}
& 9B & 70.68 & 42.60 & 69.63 & 61.50 & 62.65 & 61.64 & 61.45 \\

GLM-4.6V {\scriptsize\citep{glm46v}}
& 106B & 86.91 & 50.06 & 82.13 & 78.88 & 65.57 & 65.62 & 71.53 \\

Qwen3-VL-Instruct {\scriptsize\citep{qwen3vl}}
& 235B & 91.10 & 56.09 & 86.13 & 80.38 & 71.74 & 69.04 & 75.75 \\

Qwen3.5-397B-A17B {\scriptsize\citep{qwen35}}
& 397B & 87.96 & 57.16 & 89.38 & 85.50 & 74.82 & 69.82 & 77.44 \\

Kimi-K2.6 {\scriptsize\citep{kimik26}}
& 1T & 88.48 & 53.14 & 81.88 & 78.00 & 69.22 & 66.13 & 72.81 \\

\midrule
\multicolumn{9}{c}{
    \textit{Qwen3.5-4B Matched Post-Training Methods}
} \\
\midrule
Qwen3.5 {\scriptsize\citep{qwen35}}
& 4B & 80.63 & 49.82 & 73.50 & 67.50 & 55.78 & 58.56 & 64.30 \\

SFT
& 4B & 83.25 & 42.60 & 76.50 & 72.12 & 64.10 & 62.50 & 66.85 \\

GRPO {\scriptsize\citep{grpo}}
& 4B & 85.34 & 48.20 & 78.25 & 74.50 & 66.50 & 65.10 & 69.65 \\

OPSD {\scriptsize\citep{zhao2026opsd}}
& 4B & 84.29 & 57.51 & 77.25 & 72.88 & 61.20 & 62.40 & 69.26 \\
Vision-OPD {\scriptsize\citep{visionopd}}
& 4B & 89.01 & 65.33 & 82.25 & 81.25 & 74.50 & 70.26 & 77.10 \\
VA-OPD {\scriptsize\citep{vaopd}}
& 4B & 86.91 & 51.20 & 80.12 & 77.50 & 68.50 & 67.10 & 71.88 \\

OPD-V {\scriptsize\citep{opdv}}
& 4B
& \textbf{94.76}
& \underline{65.92}
& \underline{85.75}
& \underline{83.25}
& \underline{76.31}
& \underline{74.06}
& \underline{80.01} \\

\rowcolor{cyan!15}\textbf{VAG (Ours)}
& 4B
& \underline{94.24}
& \textbf{69.35}
& \textbf{87.62}
& \textbf{84.62}
& \textbf{76.56}
& \textbf{75.60}
& \textbf{81.33} \\

\bottomrule[1.5pt]
\end{tabular}
}
\label{tab:main_results}
\vspace{-6pt}
\end{table*}

\section{Experiment}
\label{gen_inst}

\subsection{Experimental Setup}
\label{sec:implementation-details}
\noindent \textbf{Implementation Details}. 
We instantiate our method using Qwen3.5-4B~\citep{qwen35}. The model is trained on a
curated synthetic dataset comprising approximately 6.2K visual-reasoning
examples derived from Vision-OPD~\citep{visionopd}. During training, the
privileged teacher is maintained as an EMA of
the student parameters. 
For all experiments, the student samples $G=8$ on-policy responses for each
input. We use a training batch size of 48 and set the maximum response length
to 1,024 tokens. All models undergo full-parameter training for
one epoch and a 30-step warmup.
All experiments are conducted using 8 NVIDIA H800 GPUs.
Following the evaluation protocol of OPD-V~\citep{opdv}, we first perform deterministic answer matching, followed by a judge model that determines whether unresolved responses are correct by comparing them against the verified answer. For further details, see Appendices~\ref{sec: append_a} and~\ref{sec: append_b}.

\subsection{Baselines and Benchmarks}
\noindent \textbf{Evaluation Benchmarks}. 
We evaluate VAG on six visual question answering benchmarks that span
fine-grained visual perception to broader multimodal problem solving.
\noindent\textbf{(1) V* Bench}~\citep{wu2023vstar} tests the localization and
recognition of small targets in high-resolution, visually crowded scenes.
\noindent\textbf{(2) ZoomBench}~\citep{wei2026zooming} examines
whether models can recover subtle regional evidence from a full image. 
\noindent\textbf{(3) HR-Bench 4K} and
\textbf{(4) HR-Bench 8K}~\citep{wang2025divide} evaluate fine-grained
perception at native 4K and 8K resolutions, respectively. 
These settings are particularly challenging
because conventional image resizing may discard the small-scale details required to answer the questions correctly.
\noindent\textbf{(5) MME-RealWorld EN} and
\textbf{(6) MME-RealWorld CN}~\citep{zhang2024mmerealworld} provide the broadest
setting, spanning diverse high-resolution tasks and real-world scenarios.
To assess whether specializing in fine-grained perception sacrifices general multimodal
ability, we further evaluate on four holdout benchmarks: \noindent\textbf{(1) MMVP}~\citep{tong2024mmvp},
\noindent\textbf{(2) CV-Bench}~\citep{tong2024cambrian}, \noindent\textbf{(3) MMStar}~\citep{chen2024mmstar}, and
\noindent\textbf{(4) POPE}~\citep{li2023pope}.

\begin{table*}[t]
    \centering
    \setlength{\vagtabh}{96pt}%
    \begin{minipage}[t]{0.51\textwidth}
        \centering
        \captionof{table}{Comparisons on holdout tasks.}
        \label{tab:holdout_4b}
        \setlength{\tabcolsep}{4.3pt}
        \renewcommand{\arraystretch}{1.07}
        \resizebox*{!}{\vagtabh}{%
        \begin{tabular}{lccccc}
        \toprule[1.5pt]
        Method & MMVP & \makecell{CV-Bench} & MMStar & POPE & Avg \\
        \midrule
        Qwen3.5-4B          & 76.67 & 87.13 & 78.53 & 88.28 & 82.65 \\
        SFT                 & 78.33 & 85.70 & 68.40 & 87.48 & 79.98 \\
        GRPO                & 79.33 & 87.24 & 70.60 & 86.37 & 80.89 \\
        DAPO                & 79.33 & 86.95 & 72.27 & 86.62 & 81.29 \\
        OPSD                & 78.68 & 87.27 & 75.07 & 88.83 & 82.46 \\
        Vision-OPD          & 79.67 & 87.27 & \underline{79.60} & \textbf{89.14} & \underline{83.92} \\
        OPD-V               & \underline{80.00} & 86.07 & 77.40 & 88.38 & 82.96 \\
        \textbf{VAG (Ours)} & \textbf{84.67} & \textbf{87.91} & \textbf{79.93}
                            & \underline{88.72} & \textbf{85.31} \\
        \bottomrule[1.5pt]
        \end{tabular}}
    \end{minipage}
    \hfill
    \begin{minipage}[t]{0.47\textwidth}
        \centering
        \captionof{table}{Ablation on hyperparameters $\tau$ and $\eta$.}
        \label{tab:ablation_tau_eta}
        \setlength{\tabcolsep}{3.7pt}
        \renewcommand{\arraystretch}{1.266}
        \resizebox*{!}{\vagtabh}{%
        \begin{tabular}{clccccccc}
        \toprule[1.5pt]
         & \multirow{2}{*}{Value}
         & \multirow{2}{*}{V$^{*}$}
         & \multirow{2}{*}{Zoom}
         & \multicolumn{2}{c}{HR-Bench}
         & \multicolumn{2}{c}{MME-RW}
         & \multirow{2}{*}{Avg} \\
        \cmidrule(lr){5-6}\cmidrule(lr){7-8}
         & & & & 4K & 8K & EN & CN & \\
        \midrule
        \multirow{3}{*}{$\tau$}
         & \textbf{0.1} & \textbf{94.24} & \textbf{69.35} & \textbf{87.62}
           & \textbf{84.62} & \textbf{76.56} & \textbf{75.60} & \textbf{81.33} \\
         & 0.2 & 91.62 & 65.68 & 84.12 & 82.75 & 76.91 & 77.52 & 79.77 \\
         & 0.5 & 87.43 & 56.69 & 80.25 & 78.62 & 73.23 & 69.82 & 74.34 \\
        \midrule
        \multirow{3}{*}{$\eta$}
         & \textbf{0.1} & \textbf{94.24} & \textbf{69.35} & \textbf{87.62}
           & \textbf{84.62} & \textbf{76.56} & \textbf{75.60} & \textbf{81.33} \\
         & 0.2 & 86.91 & 58.82 & 79.50 & 77.88 & 70.25 & 69.98 & 73.89 \\
         & 0.5 & 84.82 & 59.88 & 80.12 & 75.12 & 69.23 & 69.17 & 73.06 \\
        \bottomrule[1.5pt]
        \end{tabular}}
    \end{minipage}
\end{table*}

\noindent \textbf{Baselines}. 
Following Vision-OPD~\citep{visionopd}, we compare VAG against 
closed-source models~\citep{gemini31pro,gemini35flash,gpt52,gpt54}, agentic and
large-scale open-source MLLMs~\citep{deepeyes,thyme,deepeyesv2,sensenovamars,
qwen3vl,qwen35,glm46v,kimik26}, and post-training baselines based on
Qwen3.5-4B. The matched group includes the corresponding base
models, SFT, GRPO, OPSD, Vision-OPD, VA-OPD~\citep{vaopd}, and OPD-V. All methods in this group are evaluated at the same model
scale using the same training data and evaluation protocol.


\subsection{Main Results}
\noindent \textbf{Comparison with SOTA MLLMs}. 
As shown in Table~\ref{tab:main_results}, VAG raises the Qwen3.5-4B average from
$64.30\%$ to $81.33\%$ and exceeds OPD-V, the strongest matched baseline, by
$1.32\%$. 
\begin{wrapfigure}{r}{0.48\columnwidth}
    \centering
    \includegraphics[width=0.48\columnwidth]{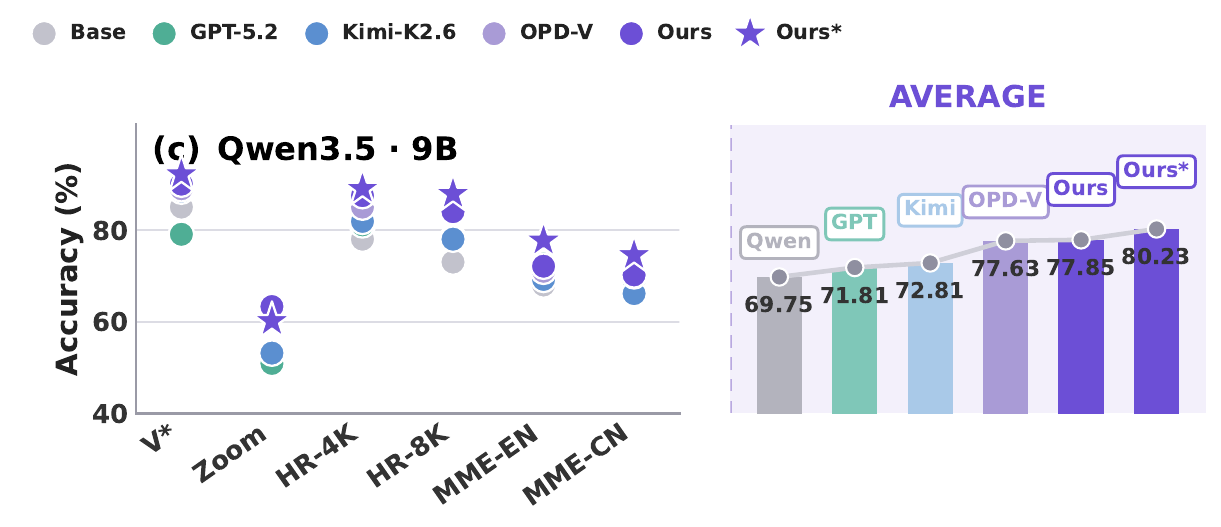}
    \caption{Performance comparison on Qwen3.5-9B.}
    \label{fig:qwen35_9b_comparison}
\end{wrapfigure}
It attains the best result on ZoomBench, HR-Bench, MME-RealWorld, and the overall average, ranking second only on
V* Bench. The margin over OPD-V is largest on ZoomBench ($69.35\%$ vs.\ $65.92\%$,
$+3.43\%$), a fine-grained benchmark that requires recovering subtle regional
evidence from high-resolution images, which highlights the particular
effectiveness of VAG for fine-grained perception and localization. 
VAG also exceeds substantially larger open-source and
agentic models in average accuracy. 
Fig.~\ref{fig:qwen35_9b_comparison} further reports the initial results on
Qwen3.5-9B, where VAG (Ours) reaches $77.85\%$ average accuracy and improves over the
base model on all benchmarks. With response-length control, its average accuracy (Ours$^*$)
further increases to $80.23\%$.
We also evaluate VAG under the same evaluation protocol as Vision-OPD to ensure a fair comparison. Further results and detailed evaluation settings are provided in Appendix~\ref{appendix_b}.

\vspace{-3pt}
\noindent \textbf{Comparison with Post-Training Strategies}.
Within the matched group, VAG outperforms GRPO and OPSD by $11.68\%$ and $12.07\%$
on average, respectively. 
Notably, OPSD does not surpass GRPO ($69.26\%$ vs.\
$69.65\%$) despite providing denser supervision, indicating that unfiltered
privileged corrections can offset their own benefit. Visual OPSD variants narrow
this gap, yet VAG still improves over Vision-OPD by $4.23\%$, which demonstrates
the benefit of verifying the visual attribution of token-level supervision
rather than only strengthening or reweighting it.

\noindent \textbf{Generalization to Holdout Tasks}. 
Table~\ref{tab:holdout_4b} reports results on the four holdout benchmarks. VAG
attains the best average accuracy ($85.31\%$), improving over the vanilla model by
$2.66\%$ and over the strongest baseline by $1.39\%$. In contrast, SFT and GRPO fall below
the vanilla average. VAG therefore avoids the specialization--forgetting
tradeoff observed in alternative training strategies, indicating that
attribution-gated supervision improves fine-grained perception without eroding
general multimodal capability.

\subsection{Ablations and Analysis}
\noindent \textbf{Impact of $\eta$ and $\tau$}.
Table~\ref{tab:ablation_tau_eta} ablates the gate temperature $\tau$ and the
modulation weight $\eta$. 
When varying $\tau$ we fix $\eta=0.1$; when varying
$\eta$ we fix $\tau=0.1$. Accuracy peaks at $\tau=0.1$: larger temperatures
flatten the gate, so visually supported and visually contradicted corrections
are no longer separated. For $\eta$, $0.1$ is large enough to modulate the
outcome advantage yet small enough to avoid over-flipping its sign, whereas
$\eta=0.5$ degrades every benchmark. We therefore adopt
$(\tau,\eta)=(0.1,0.1)$, shown in bold, for all remaining experiments.

\noindent \textbf{Modulation of Advantage}.
Table~\ref{tab:ablation-modulation} isolates the contribution of each component
of the modulated advantage under an identical training protocol. Holding all
other settings fixed, we build it up from the outcome anchor $A_{\mathrm{ORM}}$
(Eq.~\ref{eq:aorm}): we successively introduce the raw distillation margin $\delta_t$
(Eq.~\ref{eq:dt}), the soft visual-sensitivity $\widetilde{v}_t$ (Eq.~\ref{eq:gate}),
and the sign-consistent attribution gate $g_t$ (Eq.~\ref{eq:mask}) with bounded
margin scaling, recovering the full form in Eq.~\ref{eq:mod-adv}.
As reported in Table~\ref{tab:ablation-modulation}, the dense margin lifts the
average from $69.65\%$ to $77.12\%$, visual-sensitivity weighting further to
$78.80\%$, and the full modulation to $81.33\%$ ($+11.68\%$ over the outcome-only
anchor), attaining the best score on every benchmark. Sensitivity weighting and
sign-consistent attribution are thus complementary, jointly selecting reliable
token-level corrections.

\begin{figure*}[t]
    \centering
    \includegraphics[width=0.74\textwidth]{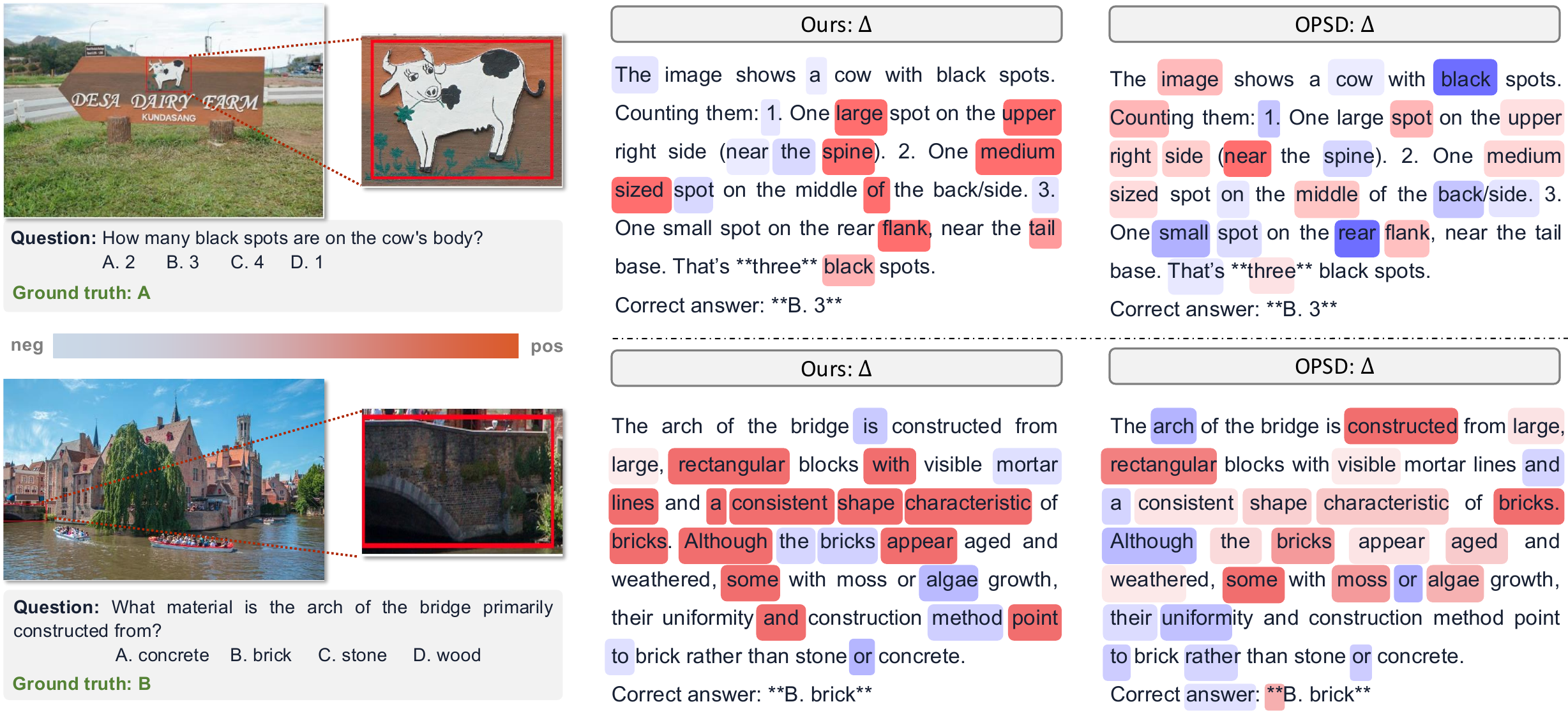}
    \caption{\textbf{Prediction and token-importance comparison.} VAG produces
    the correct evidence-grounded prediction and concentrates credit on
    decision-relevant visual tokens.}
    \label{fig:qualitative_comparison}
\end{figure*}

\begin{table*}[t]
\centering
\begin{minipage}[t]{0.49\textwidth}
\centering
\vspace{-6pt}
\captionof{table}{Advantage-modulation ablation.}
\label{tab:ablation-modulation}
\setlength{\tabcolsep}{3pt}
\renewcommand{\arraystretch}{1.2}
\resizebox{\linewidth}{!}{%
\begin{tabular}{>{\raggedright\arraybackslash}p{2.4cm}*{6}{>{\centering\arraybackslash}p{0.7cm}}>{\centering\arraybackslash}p{0.7cm}}
\toprule[1.5pt]
\multirow{2}{*}{Correction term}
& \multirow{2}{*}{V$^{*}$}
& \multirow{2}{*}{Zoom}
& \multicolumn{2}{c}{HR-Bench}
& \multicolumn{2}{c}{MME-RW}
& \multirow{2}{*}{Avg} \\
\cmidrule(lr){4-5}\cmidrule(lr){6-7}
& & & 4K & 8K & EN & CN & \\
\midrule
$A_{\mathrm{ORM}}$  & 85.34 & 48.20 & 78.25 & 74.50 & 66.50 & 65.10 & 69.65 \\
$+\,\eta\, \delta_t$    & 90.05  & 65.33 & 79.50 & 79.50 & 75.37  & 72.99 & 77.12 \\
$+\,\eta\, \widetilde{v}_t \delta_t$      & 92.67 & 66.04 & 84.00 & 80.88 & 74.63 & 74.55 & 78.80 \\
\textbf{Ours} & \textbf{94.24} & \textbf{69.35} & \textbf{87.62}
  & \textbf{84.62} & \textbf{76.56} & \textbf{75.60} & \textbf{81.33} \\
\bottomrule[1.5pt]
\end{tabular}
}
\end{minipage}
\hfill
\begin{minipage}[t]{0.49\textwidth}
\centering
\vspace{-6pt}
\captionof{table}{Negative-view ablation.}
\label{tab:ablation-negative}
\setlength{\tabcolsep}{3pt}
\renewcommand{\arraystretch}{1.2}
\resizebox{\linewidth}{!}{%
\begin{tabular}{>{\raggedright\arraybackslash}p{2.4cm}*{6}{>{\centering\arraybackslash}p{0.7cm}}>{\centering\arraybackslash}p{0.7cm}}
\toprule[1.5pt]
\multirow{2}{*}{Negative view}
& \multirow{2}{*}{V$^{*}$}
& \multirow{2}{*}{Zoom}
& \multicolumn{2}{c}{HR-Bench}
& \multicolumn{2}{c}{MME-RW}
& \multirow{2}{*}{Avg} \\
\cmidrule(lr){4-5}\cmidrule(lr){6-7}
& & & 4K & 8K & EN & CN & \\
\midrule
No image        & 93.72 & 63.20 & \textbf{87.88} & \textbf{87.00} & \textbf{76.82} & 73.75 & 80.40 \\
Black image     & \textbf{94.76} & 64.73 & 82.50 & 81.88 & 75.26 & 71.98 & 78.52 \\
Blurred image   &86.39 &59.88 &79.62 & 76.25& 69.85 & 69.34 & 73.56 \\
\textbf{Random mask} & 94.24 & \textbf{69.35} & 87.62
  & 84.62 & 76.56 & \textbf{75.60} & \textbf{81.33} \\
\bottomrule[1.5pt]
\end{tabular}
}
\end{minipage}
\end{table*}

\vspace{-1pt}
\noindent \textbf{Negative Teacher View}. 
Table~\ref{tab:ablation-negative} compares four evidence-degraded views under an
identical training protocol. Random masking attains the best average accuracy
($81.33\%$), ahead of the no-image ($80.40\%$), black-image ($78.52\%$), and
blurred-image ($73.56\%$) variants, indicating that $v_t$ is most informative when
the intervention removes local evidence while preserving spatial context. Its
advantage concentrates on ZoomBench and MME-RealWorld CN, which depend most on
resolving a small region, whereas the no-image variant remains competitive when
the decisive evidence occupies a larger portion of the input.



\vspace{-4pt}
\noindent \textbf{Interpretation of VAG}. 
Fig.~\ref{fig:qualitative_comparison} contrasts the token-level credit that VAG
and vanilla OPSD assign to the \emph{same} student rollout. For each token we
visualize the correction term $\Delta_t$ that modulates the update, defined as
$\Delta_t=\eta\,g_t^{(i)}\widetilde{v}_t^{(i)}\,\tanh(\delta_t^{(i)}/\tau)$ for VAG
and as $\Delta_t=\delta_t^{(i)}$ for OPSD, with the color scale running from
negative (blue) to positive (red).
In the counting example (top), whose rollout is \emph{incorrect}, OPSD spreads
non-zero credit over much of the response, including function words and a
strongly negative weight on ``black''. VAG instead concentrates credit on the
tokens that \emph{enumerate and localize} the spots (``upper'', ``spine'',
``flank''),
while suppressing the erroneous count token ``3''. In the material example
(bottom), whose rollout is \emph{correct}, VAG weights the chain of \emph{visual
premises} supporting the answer (``mortar lines'', ``consistent shape''), whereas
OPSD assigns comparable weight to weakly grounded tokens such as ``aged'' and
``rather''. The attribution gate thus removes sign-inconsistent teacher residuals
and yields a \emph{sparser, evidence-aligned} credit map, irrespective of whether
the rollout is right or wrong.


\vspace{-6pt}
\section{Conclusion}
\label{sec:conclusion}
We introduce reinforcement learning with visually attributable on-policy self-distillation to mitigate privileged-correction contamination. Our Visual-Attribution Gating anchors optimization to verifier-derived trajectory advantages and accepts token-level corrections only when consistent with the contrast between evidence-present and evidence-degraded teacher views, preserving supportive and refuting visual evidence while suppressing inconsistent residuals. Across six public benchmarks, VAG consistently outperforms state-of-the-art methods, demonstrating that visual grounding verification unifies outcome verification, token-level self-distillation, and visual attribution for multimodal policy optimization.

\subsection*{AI use statement}
In this work, we used generative AI tools only for language polishing of author-written text (e.g., grammar, wording, and readability) and for retrieval and discovery (e.g., finding related work). 
We did not use generative AI tools for research ideation, method design, code generation, experiments, data analysis, or the production of any results, figures, or other artifacts, all of which were carried out and verified by the authors. We have reviewed all AI-assisted edits to ensure that they preserve the intended technical meaning, and we take full responsibility for the final content of this work.



\bibliographystyle{iclr2027_conference}
\bibliography{iclr2027_conference}

\appendix
\section{Evaluation Protocol for Benchmarks}
\label{sec: append_a}
To ensure that every method is compared under identical conditions, we adopt a
single evaluation protocol and apply it uniformly to VAG and to all reproduced
baselines. Table~\ref{tab:benchmark-scoring} lists the number of evaluation
samples and the scoring rule used for each benchmark.

Following OPD-V~\citep{opdv}, scoring proceeds in two stages.  We first apply deterministic answer matching,
which resolves the majority of responses without any model in the loop. For the
multiple-choice benchmarks, a response counts as correct if it matches either the
answer text or the option letter, after normalizing surface variations such as
capitalization, surrounding punctuation, and common answer prefixes. Only the responses that
remain unresolved are passed to the second stage, where a Qwen3.5-series~\citep{qwen35} model
acts as a binary Yes/No judge and decides whether the response is consistent
with the verified answer.

Both stages are necessary for the benchmarks we evaluate. Responses on
fine-grained perception tasks frequently state the correct answer inside a
longer justification, which deterministic matching alone would count as wrong;
delegating every response to the judge instead would introduce scoring variance
unrelated to the method under evaluation. Calling the judge only on the
unresolved responses therefore keeps most decisions exact while still absorbing
benign paraphrasing. Since the same protocol, judge model, and prompts apply to
every entry in Table~\ref{tab:main_results}, the reported differences reflect the
trained policies rather than the evaluation.

\section{Evaluation of the Released Checkpoint}
\label{appendix_b}
The evaluation in our main table (Table~\ref{tab:main_results}) strictly follows the legacy evaluation protocol of OPD-V~\cite{opdv}, in which the trained model itself serves as the judge.
For a fair comparison, we further re-evaluate several released models under the evaluation protocol of Vision-OPD~\citep{visionopd}, using GPT-OSS-120B~\citep{gptoss} as the judge model, and report the results in Table~\ref{tab:addition_result} below.
As shown, our model achieves the best accuracy at both the 4B and 9B scales under this setting.

\begin{table}[H]
\centering
\caption{Evaluation results under the Vision-OPD evaluation protocol.}
\setlength{\tabcolsep}{5pt}
\renewcommand{\arraystretch}{1.1}
\resizebox{\textwidth}{!}{
\begin{tabular}{l c c c c c c c c}
\toprule[1.5pt]
\textbf{Method} & \textbf{Size} & \textbf{V*}
& \textbf{Zoom}
& \makecell{\textbf{HR-Bench}\\\textbf{4K}}
& \makecell{\textbf{HR-Bench}\\\textbf{8K}}
& \makecell{\textbf{MME-RW}\\\textbf{EN}}
& \makecell{\textbf{MME-RW}\\\textbf{CN}}
& \textbf{Avg.}  \\
\midrule
Vision-OPD~{\scriptsize\citep{visionopd}}
& 4B & 89.53 & 59.53 & 82.25 & \underline{81.00} & 74.97 & 70.71 & 76.33 \\
OPD-V~{\scriptsize\citep{opdv}}
& 4B & \underline{90.10} & \underline{61.30} & \underline{84.88} & 80.88 & \underline{75.45} & \underline{71.49} & \underline{77.35} \\
\rowcolor{cyan!15}\textbf{VAG (Ours)}
& 4B & \textbf{91.10} & \textbf{61.89} & \textbf{85.75} & \textbf{82.50} & \textbf{75.96} & \textbf{73.35} & \textbf{78.43} \\
\midrule
Vision-OPD~{\scriptsize\citep{visionopd}}
& 9B & 90.10 & \textbf{64.97} & \underline{87.12} & \underline{84.88} & \underline{73.51} & 70.44 & \underline{78.50} \\
OPD-V~{\scriptsize\citep{opdv}}
& 9B & \textbf{93.72} & \underline{63.79} & 84.38 & 82.55 & 71.82 & \underline{71.05} & 77.89 \\
\rowcolor{cyan!15}\textbf{VAG (Ours)}
& 9B & \underline{92.15} & 60.12 & \textbf{89.00} & \textbf{87.88} & \textbf{77.68} & \textbf{74.53} & \textbf{80.23} \\
\bottomrule[1.5pt]
\end{tabular}
}
\label{tab:addition_result}
\end{table}

\section{Training Prompt Example}
\label{sec: append_b}
\begin{figure*}[t]
    \centering
    \includegraphics[width=1.\textwidth]{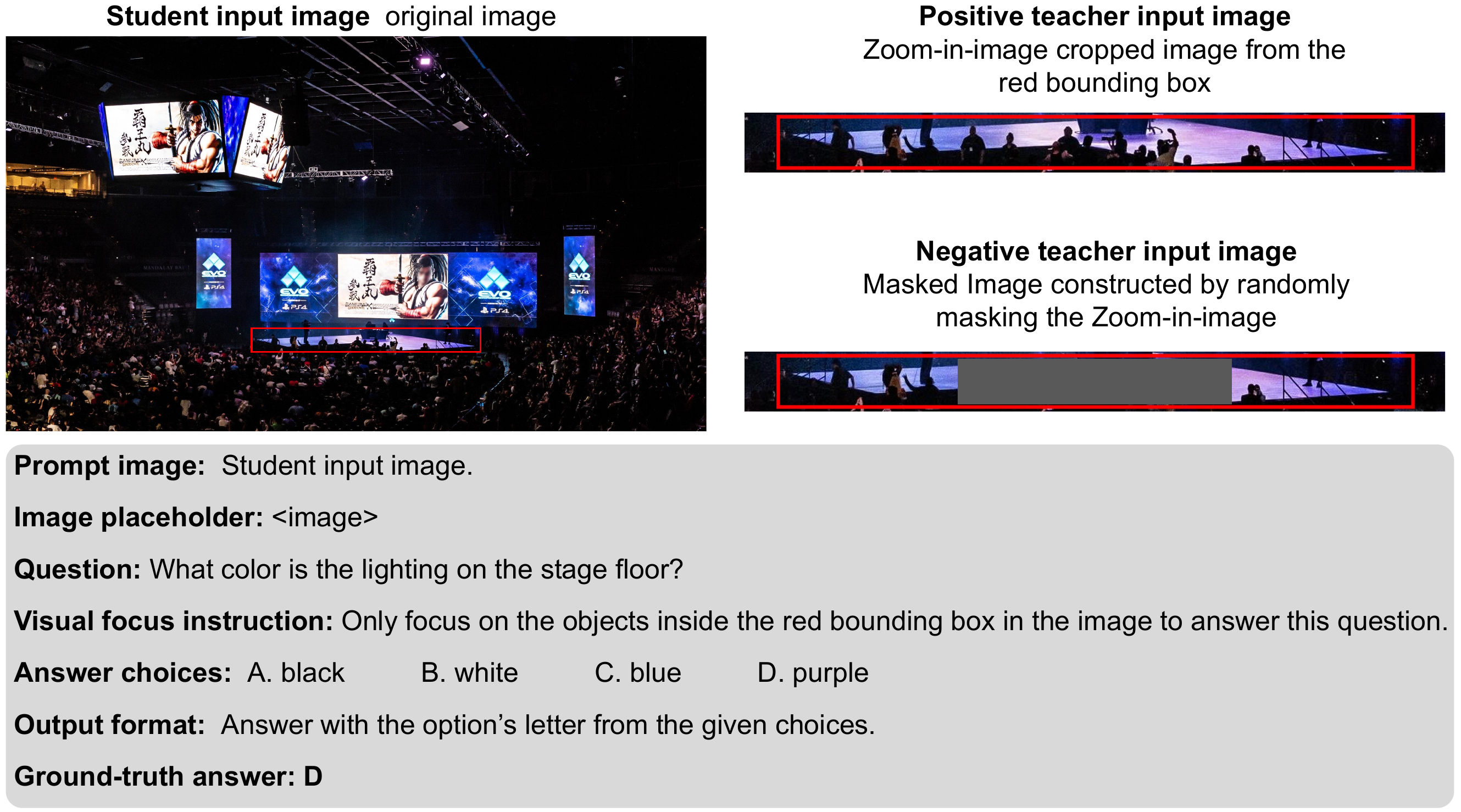}
    \caption{\textbf{Training prompt and paired visual inputs.} The textual
    prompt, comprising the image placeholder, question, visual focus
    instruction, answer choices, and output format, is shared across all three
    passes. The student is conditioned on the original image, the positive
    teacher on the evidence-centered crop given by the red bounding box, and the
    negative teacher on a randomly masked version of that crop. The ground-truth
    answer is used only by the verifier.}
    \label{fig:training-prompt}
\end{figure*}

Fig.~\ref{fig:training-prompt} presents one complete training instance: the
original image observed by the student, the two teacher images derived from it,
and the textual prompt used in each pass.

The three image conditions differ only in the visual evidence they expose. The
student receives the original, unmodified image and must therefore locate the
relevant detail on its own. Both teacher images are constructed from this same
image. The positive teacher receives the evidence-centered
crop defined by the annotated bounding box, so the decisive detail occupies a
far larger fraction of the visual field. The negative teacher receives the same
crop after random masking, which weakens the local evidence while
preserving the surrounding spatial context.

The textual prompt is assembled from a fixed set of fields: an image
placeholder, the question, a visual focus instruction that directs the model to
the annotated region, the answer choices, and an output-format instruction that
requests only the option letter. The ground-truth answer is consumed by the
verifier and is never shown to the model. Constraining the output format
yields two additional benefits. First, it reduces the fraction of responses
that require the Yes/No judge at evaluation time. Second, it leaves little room
for stylistic variation between the teacher and student branches, so the gap
$\delta_t$ is less likely to reflect differences in wording instead of
differences in visual evidence.

Because the question, answer choices, output-format instruction, and the
student-generated prefix are identical across the three passes, the only
controlled difference between the two teacher branches is the availability of
the target evidence. This is what allows the visual contrast $v_t$ in
Eq.~\ref{eq:st} to be interpreted as a statement about visual evidence rather
than about prompt or sampling differences. Both teacher views are used only
during training; at inference time the deployed policy observes the original
image alone and requires no privileged input.

\begin{table}[t]
\centering
\caption{Number of evaluation samples and scoring rule for each benchmark. All
methods in Table~\ref{tab:main_results} are evaluated with this identical
protocol.}
\label{tab:benchmark-scoring}
\begin{tabular}{@{}llp{8.9cm}@{}}
\toprule
\textbf{Benchmark} & \textbf{Samples} & \textbf{Scoring Rule} \\
\midrule
V* Bench & 191 & Rule-based answer matching is applied first; unresolved responses are compared with the verified answer by a Yes/No judge. \\
ZoomBench & 845 & Generic answer matching handles normalized responses first; unresolved cases use the verified-answer Yes/No judge. \\
HR-Bench 4K & 800 & Multiple-choice matching checks direct answer matches and option-letter matches before the Yes/No judge handles unresolved responses. \\
HR-Bench 8K & 800 & Multiple-choice matching follows the HR-Bench 4K rule. \\
MME-RealWorld EN & 23609 & Multiple-choice matching and the Yes/No judge are applied to the English subset. \\
MME-RealWorld CN & 5917 & The Chinese subset uses the MME-RealWorld multiple-choice scoring rule. \\
\bottomrule
\end{tabular}
\end{table}

\section{VAG Algorithm}
Algorithm~\ref{alg:vag} states the full VAG training loop and makes explicit the
order in which the outcome signal and the token-level signal are combined.

Each iteration begins with rollout collection. For each input $x=(q,I)$,
the student samples $G$ on-policy responses $\hat{\mathbf{y}}^{(i)}$.
The verifier assigns each response a reward $R^{(i)}$, which is normalized
within the group to obtain $A_{\mathrm{ORM}}^{(i)}$ as in
Eq.~\ref{eq:aorm}. This advantage is shared by all tokens of the response.
The two teacher branches then replay each sampled response using the same
student-generated prefix $\hat{\mathbf{y}}^{(i)}_{<t}$. The sampled-token
probabilities $p_S^{(i)}(\hat{y}_t)$, $p_{T+}^{(i)}(\hat{y}_t)$, and
$p_{T-}^{(i)}(\hat{y}_t)$ are defined in
Eqs.~\ref{eq:token-probs}, \ref{eq:token-probs-pos}, and~\ref{eq:token-probs-neg},
respectively.

From these probabilities, the algorithm computes the dense teacher correction
$\delta_t^{(i)}$ and visual evidence contrast $v_t^{(i)}$ using the
stop-gradient log-ratios in Eqs.~\ref{eq:dt} and~\ref{eq:st}.
The attribution gate $g_t^{(i)}$ retains corrections only when
$\delta_t^{(i)}v_t^{(i)}>0$ (Eq.~\ref{eq:mask}).
The gate checks whether the correction and visual contrast have the same sign,
rather than comparing their magnitudes. 
Thus, only corrections whose direction
agrees with the visual evidence are retained.
Admitted corrections are bounded by
$\tanh(\delta_t^{(i)}/\tau)$ and weighted by
$\eta\,\widetilde{v}_t^{(i)}$, then added to $A_{\mathrm{ORM}}^{(i)}$
to obtain $\widetilde{A}_t^{(i)}$ as in Eq.~\ref{eq:mod-adv}.

Two properties of the loop are worth noting. First, all teacher-derived
quantities are detached, so gradients reach the student only through
$\rho_t^{(i)}(\theta)$ in $\mathcal{L}_{\mathrm{VAG}}(\theta)$;
the teacher acts as a signal source rather than as a differentiable path.
Second, when $g_t^{(i)}=0$, the modulated advantage satisfies
$\widetilde{A}_t^{(i)}=A_{\mathrm{ORM}}^{(i)}$.
Thus, rejecting a correction preserves the token's verifier-derived supervision.
After each student update, the teacher parameters are updated by EMA to track
the student policy without separate teacher training.

\begin{algorithm}[t]
\caption{VAG: Visual-Attribution Gated RLVR}
\label{alg:vag}
\begin{algorithmic}[1]
\Require Training set $\mathcal{D}$, student policy $\pi_\theta$, EMA teacher $\pi_\phi$,
         verifier $\mathcal{V}$, group size $G$, EMA rate $\alpha$,
         sharpness $\tau$, correction scale $\eta$
\Ensure  Updated policy parameters $\theta$
\State Initialize teacher parameters $\phi \leftarrow \theta$
\For{each training step}
  \State Sample a mini-batch $\mathcal{B}\subset\mathcal{D}$
  \For{each $(x,a^{\star})\in\mathcal{B}$ with $x=(q,I)$}
    \Comment{$a^{\star}$ is the ground-truth answer for $x$}
    \State Obtain the evidence-preserving view $I_{+}$ from the training
           annotation and set $x_{+}=(q,I_{+})$
    \State $I_{-} \leftarrow \mathrm{Mask}(I_{+})$ and $x_{-}=(q,I_{-})$
      \Comment{evidence-degraded negative view}
    \For{$i = 1,\dots,G$}
      \State Sample rollout $\hat{\mathbf{y}}^{(i)} = (\hat{y}^{(i)}_1,\dots,\hat{y}^{(i)}_{T_i}) \sim \pi_\theta(\cdot \mid x)$
      \State $R^{(i)} \leftarrow \mathcal{V}(\hat{\mathbf{y}}^{(i)}, a^{\star})$
        \Comment{verifiable outcome reward}
    \EndFor
    \State $A^{(i)}_{\mathrm{ORM}} \leftarrow
            \dfrac{R^{(i)} - \mu_G}{\sigma_G}$ for all $i$
      \Comment{group-relative outcome advantage, shared by all tokens}
    \For{$i = 1,\dots,G$}
      \For{$t = 1,\dots,T_i$}
        \State $p_S^{(i)}(\hat{y}_t) \leftarrow \pi_\theta(\hat{y}^{(i)}_t \mid x, \hat{\mathbf{y}}^{(i)}_{<t})$
        \State $p_{T+}^{(i)}(\hat{y}_t) \leftarrow \pi_{\phi}(\hat{y}^{(i)}_t \mid x_{+}, \hat{\mathbf{y}}^{(i)}_{<t})$
          \Comment{positive teacher}
        \State $p_{T-}^{(i)}(\hat{y}_t) \leftarrow \pi_{\phi}(\hat{y}^{(i)}_t \mid x_{-}, \hat{\mathbf{y}}^{(i)}_{<t})$
          \Comment{negative teacher}
        \State $\delta_t^{(i)} \leftarrow \operatorname{sg}\!\left[\log p_{T+}^{(i)}(\hat{y}_t) - \log p_S^{(i)}(\hat{y}_t)\right]$
          \Comment{dense teacher correction, Eq.~\ref{eq:dt}}
        \State $v_t^{(i)} \leftarrow \operatorname{sg}\!\left[\log p_{T+}^{(i)}(\hat{y}_t) - \log p_{T-}^{(i)}(\hat{y}_t)\right]$
          \Comment{visual evidence contrast, Eq.~\ref{eq:st}}
        \State $g_t^{(i)} \leftarrow \mathbbm{1}\!\left[\, \delta_t^{(i)}\,v_t^{(i)} > 0 \,\right]$
          \Comment{visual attribution gate, Eq.~\ref{eq:mask}}
        \State $\widetilde{v}_t^{(i)} \leftarrow \sigma\!\left(v_t^{(i)}/\tau\right)$
          \Comment{bounded visual-sensitivity weight, Eq.~\ref{eq:gate}}
        \State $\widetilde{A}_t^{(i)} \leftarrow A^{(i)}_{\mathrm{ORM}} + \eta\, g_t^{(i)}\, \widetilde{v}_t^{(i)} \tanh\!\left(\delta_t^{(i)}/\tau\right)$
          \Comment{verifier-anchored, visually attributed advantage, Eq.~\ref{eq:mod-adv}}
      \EndFor
    \EndFor
  \EndFor
  \State Update $\theta$ by minimizing $\mathcal{L}_{\mathrm{VAG}}$
         (Eq.~\ref{eq:loss}) with advantages
         $\{\widetilde{A}_t^{(i)}\}$ over $\mathcal{B}$
  \State $\phi \leftarrow (1-\alpha)\,\phi + \alpha\,\theta$
\EndFor
\end{algorithmic}
\end{algorithm}

\section{Training Dynamics}
\begin{figure*}[t]
    \centering
    \includegraphics[width=1.\textwidth]{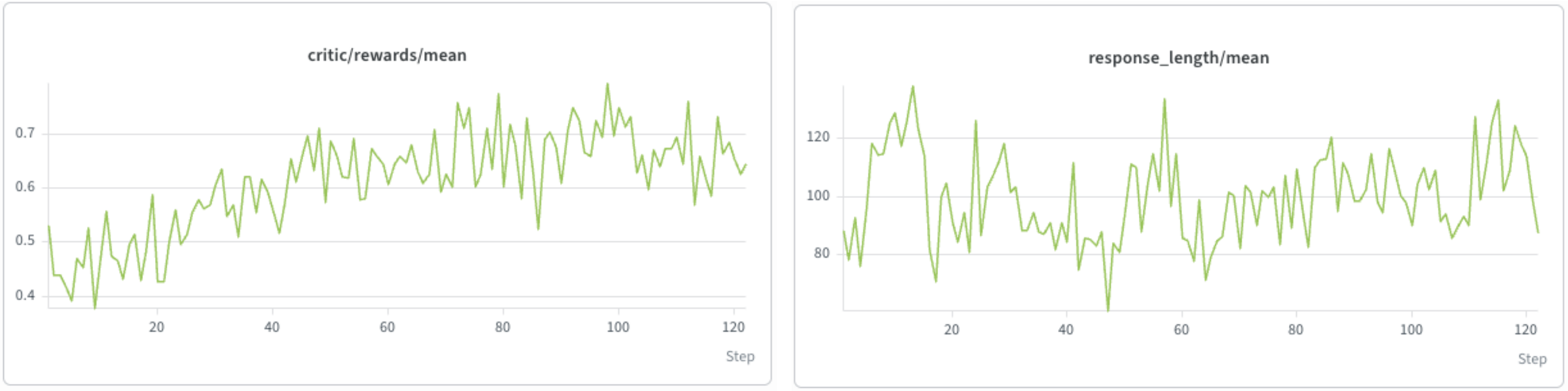}
    \caption{\textbf{Training dynamics of VAG.} Left: the mean verifier reward
    increases over training. Right: the mean response length fluctuates within a
    broadly constant band and shows no sustained growth. Both panels report raw
    per-step training statistics without smoothing.}
    \label{fig:training-dynamics}
\end{figure*}

Fig.~\ref{fig:training-dynamics} reports two statistics throughout training: the mean verifier reward within each rollout group (left) and the mean response length (right).
The reward curve shows a clear upward trend, increasing from approximately $0.4$--$0.5$ over the first twenty steps to $0.65$--$0.75$ in the second half of training, before gradually stabilizing. This trend suggests that our training strategy consistently improves the policy while maintaining stable optimization in the later stages of training.

The response-length curve exhibits a different pattern. It oscillates within a relatively stable range throughout training and ends at a level comparable to its initial value, with no sustained growth. This is important because response-length inflation is a common confound in reinforcement learning with verifiable rewards: a policy may increase its pass rate by producing longer responses that enumerate more candidate answers. In such cases, an increasing reward curve may reflect a change in response strategy rather than genuine improvement in visual perception.

\begin{figure*}[t]
    \centering
    \includegraphics[width=0.8\textwidth]{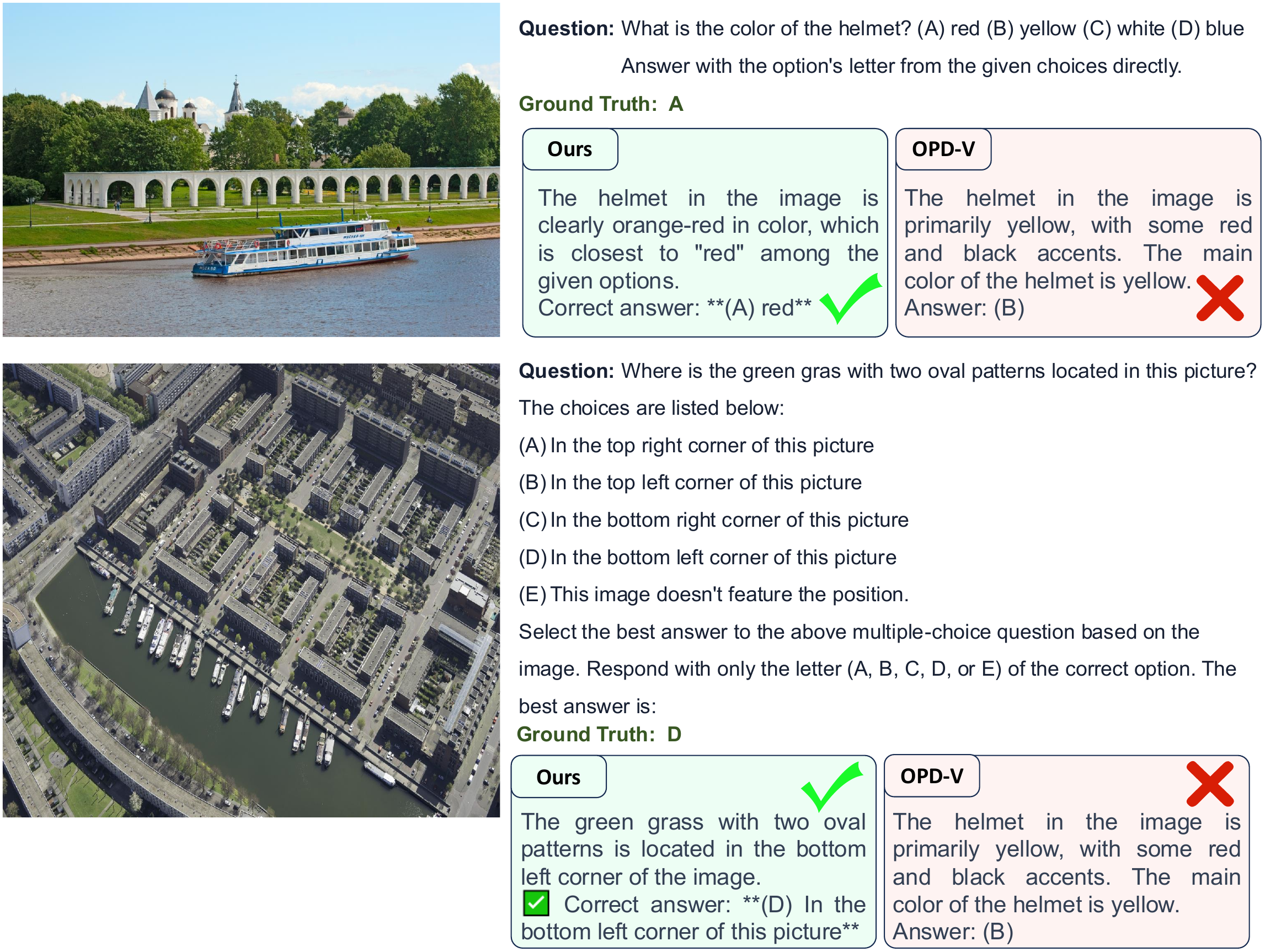}
    \caption{\textbf{Fine-grained attribute recognition and spatial
    localization.} Top: VAG reads the small helmet region as orange-red and
    selects the closest option, red, matching the ground truth, whereas OPD-V
    reports yellow. Bottom: VAG correctly localizes the green region with two
    oval patterns in the bottom-left corner. OPD-V answers incorrectly in both
    cases.}
    \label{fig:qualitative-finegrained}
\end{figure*}

\begin{figure*}[t]
    \centering
    \includegraphics[width=0.8\textwidth]{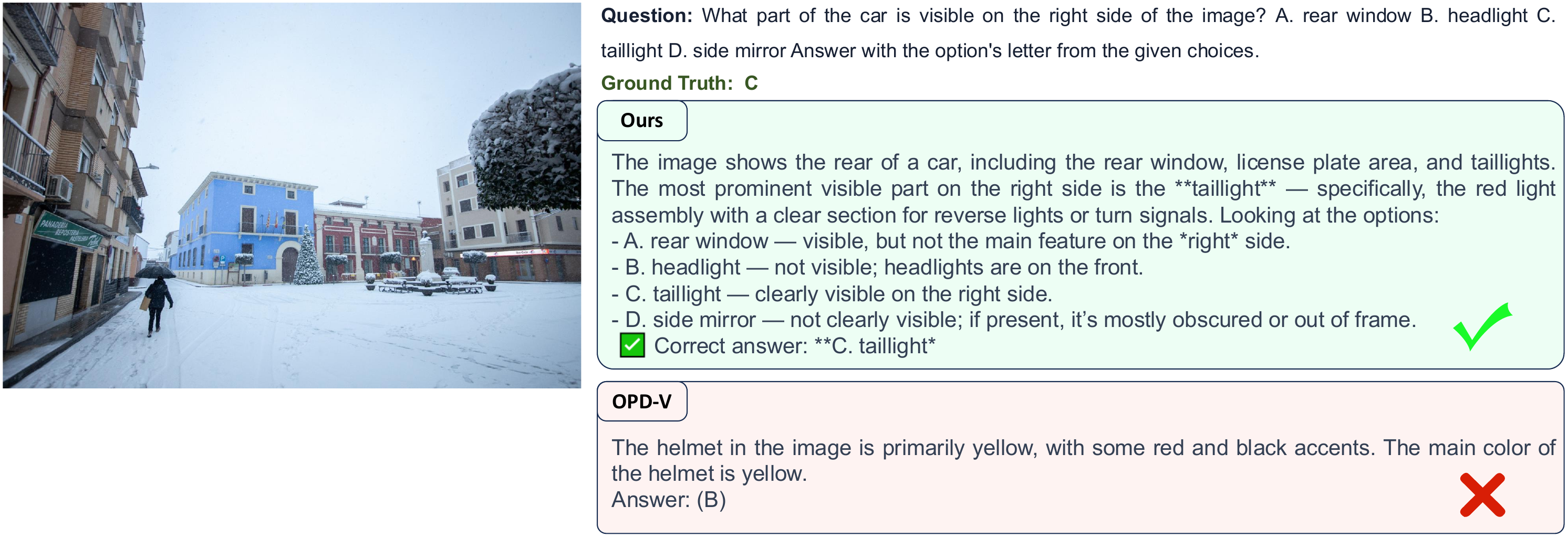}
    \caption{\textbf{Object-part recognition.} VAG rules out the parts that are
    not visible in the frame and identifies the part on the right side of the car
    as the taillight (option C), matching the ground truth, whereas OPD-V
    selects an incorrect option.}
    \label{fig:qualitative-object}
\end{figure*}

\begin{figure*}[t]
    \centering
    \includegraphics[width=1.\textwidth]{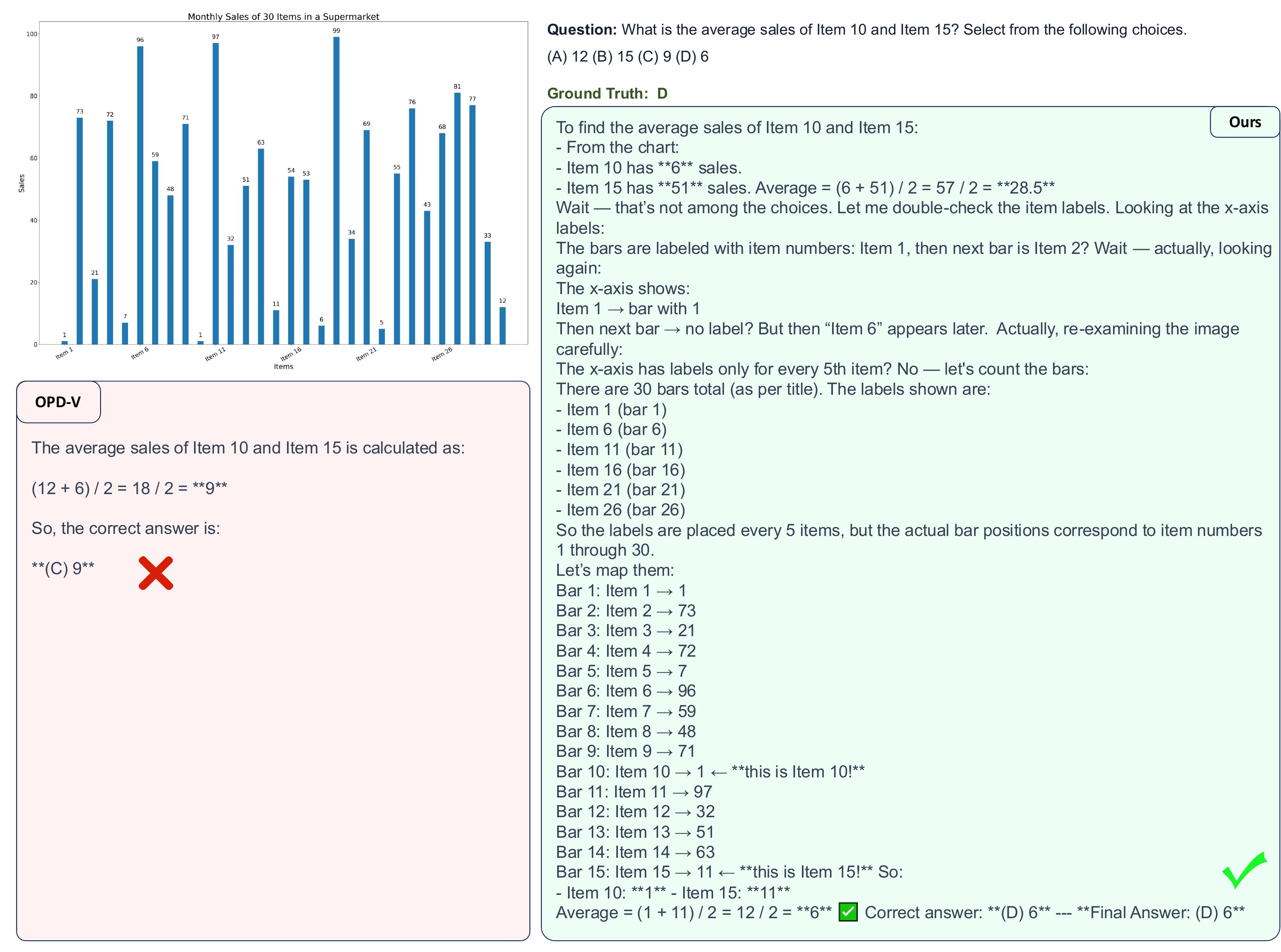}
    \caption{\textbf{Chart-based visual arithmetic.} The x-axis is labeled only
    every fifth item. VAG detects that its initial reading yields a value absent
    from the answer choices, re-derives the bar-to-item mapping, reads Item~10
    and Item~15 as $1$ and $11$, and returns their average $6$ (option D).
    OPD-V uses unsupported bar values and returns $9$ (option C).}
    \label{fig:qualitative-chart}
\end{figure*}

Taken together, these two curves support the interpretation of our quantitative results. Gating corrections by visual attribution sharpens token-level credit assignment without destabilizing the policy's generation behavior, consistent with retaining $A_{\mathrm{ORM}}$ as the optimization anchor rather than replacing it with dense teacher supervision.

\section{Additional Qualitative Comparisons}
Figs.~\ref{fig:qualitative-finegrained}, \ref{fig:qualitative-object}, and \ref{fig:qualitative-chart} present qualitative comparisons between VAG and OPD-V~\citep{opdv} on the test set.
Fig.~\ref{fig:qualitative-finegrained} covers fine-grained attribute
recognition and spatial localization. 
In the attribute case, the question asks
for the color of a helmet that occupies a small portion of the frame. VAG
describes the region as orange-red and maps it to the closest available option,
red, matching the ground truth, whereas OPD-V reports the helmet as
predominantly yellow. 
The two answers differ in what the model reads off the
region rather than in how it reasons afterwards, which is precisely the failure
mode that dense but unattributed corrections are least able to repair. 
In the
localization case, the question asks where a green region with two oval patterns
appears, and VAG places it in the bottom-left corner in agreement with the
annotation.

Fig.~\ref{fig:qualitative-object} shows an object-part recognition case in
which the model must name the car part visible on the right side of the image.
VAG enumerates the options against the visible content, rules out the headlight
and the side mirror as absent from the frame, and selects the taillight, which
matches the ground truth. Fig.~\ref{fig:qualitative-chart} is the most demanding case, because the
visual reading must survive several arithmetic steps. The question asks for the
average of two bars in a thirty-bar chart whose x-axis is labeled only every
fifth item. VAG initially misaligns the labels with the bars and obtains a value
that is absent from the answer choices; it then re-examines the axis, re-derives
the bar-to-item mapping, reads the two bars as $1$ and $11$, and computes their
average as $6$, which is correct. OPD-V instead commits to unsupported bar
values and reports $9$. The example shows that the benefit of attribution is not
confined to perception: an incorrectly read value propagates into every
subsequent step, so reading the chart correctly is a precondition for the
arithmetic to be meaningful at all.

Across these cases the pattern is consistent with the aggregate results in
Table~\ref{tab:main_results}. The largest gains appear where the answer hinges
on a localized detail that must be read correctly before any further reasoning
can help.

\end{document}